\documentclass{article}
\usepackage{spconf,amsmath,graphicx,amsfonts, hyperref}
\usepackage{xcolor}
\usepackage{multirow}
\usepackage{booktabs}
\usepackage{pifont}
\usepackage{makecell}

\newcommand{\rescell}[2]{\makecell{$#1{\scriptstyle \pm#2}$}}
\newcommand{\cmark}{\ding{51}}%
\newcommand{\xmark}{\ding{55}}

\title{Using Emotion Embeddings to transfer knowledge \\between emotions, languages, and annotation formats}

\name{\makecell{Georgios Chochlakis$^{1, 2}$\quad Gireesh Mahajan$^3$ \quad Sabyasachee Baruah$^{1, 2}$\\ Keith Burghardt$^2$ \qquad Kristina Lerman$^2$\qquad Shrikanth Narayanan$^{1, 2}$}}
\address{\normalsize $^1$ Signal Analysis and Interpretation Lab, University of Southern California, Los Angeles, CA 90089, USA \\ \normalsize $^2$ Information Science Institute, University of Southern California, Marina del Rey, CA 90292, USA \\ \normalsize $^3$ Microsoft Cognitive Services, Redmond, WA 98052, USA}

\begin{document}
\ninept
\maketitle
\begin{abstract}
The need for emotional inference from text continues to diversify as more and more disciplines integrate emotions into their theories and applications. These needs include inferring different emotion types, handling multiple languages, and different annotation formats. 
A shared model between different configurations would enable the sharing of knowledge and a decrease in training costs, and would  simplify the process of deploying emotion recognition models in novel environments. In this work, we study how we can build a single model that can transition between these different configurations by leveraging multilingual models and \textit{Demux}, a transformer-based model whose input includes the emotions of interest, enabling us to dynamically change the  emotions predicted by the model. \textit{Demux} also produces emotion embeddings, and performing operations on them allows us to transition to clusters of emotions by pooling the embeddings of each cluster. We show that \textit{Demux} can simultaneously transfer knowledge in a zero-shot manner to a new language, to a novel annotation format and to unseen emotions. Code is available at \url{https://github.com/gchochla/Demux-MEmo}.\footnote{Funded in part by DARPA under contract HR001121C0168}
\end{abstract}


\vspace{-0.5cm}
\section{Introduction}
\label{sec:intro}


Human experience is permeated by emotions. They can guide our attention and influence our information consumption, beliefs, and our interactions~\cite{dukes2021rise, wahl2019emotions}. Deep learning has enabled us to extract affective constructs from natural language~\cite{chochlakis2022leveraging, calvo2015oxford}, allowing emotion recognition from text at scale~\cite{guo2022emotion}. 
Nevertheless, the need for better performance across various metrics of interest still exists.


When inferring emotions from text, earlier approaches have utilized  emotion lexicons \cite{pennebaker2001linguistic}. These struggle in more realistic settings, because, for example, they do not handle context, like negation. On the other hand, while modern efforts relying on deep learning achieve better performance \cite{chochlakis2022leveraging, baziotis2018ntua}, these data-driven models have to contend with a multitude of biases, such as annotation biases in the data used to train the models.

The needs for emotional inference from text have also diversified. First, the domain of interest can vary greatly between applications, ranging from everyday dialogues \cite{li2017dailydialog} to tweets \cite{mohammad2018semeval}. Secondly, it is desirable for the models to be able to handle multiple languages \cite{mohammad2018semeval}, such as when studying perceptions and reactions to international news stories. Finally, the emotions of interest can differ between applications, and perhaps even the annotation scheme might not be similar. For example, in this work, we analyzed tweets annotated for clusters of emotions, where emotions that co-occur frequently were grouped, in contrast to single emotions in other settings. Hence, transfer learning is hindered by the mismatch.

In this work, our main goal is to achieve transfer of emotion recognition between annotation formats, emotions and languages. Our experimental design examines this step-by-step. First, we leverage pretrained multilingual language models \cite{barbieri2022xlmt, devlin2018bert} to enable knowledge transfer between languages. We also use and extend \textit{Demux} \cite{chochlakis2022leveraging}, a model that incorporates the labels in its input space to achieve the final classification. Emotions are then embedded in the same space as the language tokens. Emotion word embeddings can facilitate transfer between emotions, as shown in \cite{chochlakis2022leveraging}, so we study whether this can also be achieved in a zero-shot manner. Lastly, we examine how an extension of \textit{Demux} to clusters can transfer knowledge between different annotation formats by directly performing operations on label embeddings \cite{mikolov2013linguistic}. Our contributions include the following:
\begin{itemize}
    \item We show that multilingual emotion recognition models can be competitive with or even outperform monolingual baselines, and that knowledge can be transferred to new languages.
    \item We demonstrate that \textit{Demux} can inherently transfer knowledge to emotions it has not been trained with.
    \item We illustrate that operations on the contextual emotion embeddings of \textit{Demux} can successfully achieve transfer to novel annotation formats in a zero-shot manner. To the best of our knowledge, we are the first to study this setting.
    \item We show that \textit{Demux} can be critical for flexible emotion recognition in a dynamic environment with ever-changing inference needs, such as the addition and subtraction of emotion types, changes in language, and alterations in the annotation format, e.g., the clustering of different emotions.
\end{itemize}
%


\section{Related Work}
\label{sec:related}

\subsection{Emotion Recognition}

Earlier works utilized Bag-of-Words algorithms driven by emotion lexicons. For instance, LIWC \cite{pennebaker2001linguistic} is a lexicon that is widely used to perform word counting, while DDR \cite{garten2018dictionaries} extends lexicon-based methods from word counting to computing similarities between words. More recently, deep learning has enabled more accurate extraction of emotion signals from text. Initial efforts have treated the task as single-label, and use a threshold to transform into the desired multi-label output \cite{he2018joint}. LSTMs have been widely used for the task \cite{felbo2017using, baziotis2018ntua} e.g., for SemEval 2018 Task 1 \cite{mohammad2018semeval}, where some also used features from affective lexicons. More recently, Transformers \cite{vaswani2017attention} have dominated the field. \textit{Demux} and \verb+MEmo+ \cite{chochlakis2022leveraging}, state of the art models in SemEval 2018 Task 1 E-c, prompt BERT-based \cite{devlin2018bert} models in different ways, by including all emotions in the input or \verb+[MASK]+ tokens in language prompts, respectively. They also employ an intra-group correlation loss to further improve performance. Transformers have also been used with other architectures~\cite{xu2020emograph}.

\subsection{Multilingual Models \& Emotion Recognition} \label{sec:multilingual-related}

Multilingual transformers attempt to model many languages simultaneously. Normalized sampling from each language is used so that low-resource languages are not severely hindered, which we also adopt. This is achieved, given $\alpha \in [0, 1]$, by transforming the frequency $p_l$ of each language as $p_l^\prime \leftarrow p_l^{\alpha}$ and renormalizing to create the new sampling distribution \cite{lample2019cross}. Note that as $\alpha$ decreases, the distribution becomes more balanced, achieving parity at $\alpha=0$.

BERT~\cite{devlin2018bert} and XLM~\cite{lample2019cross} require preprocessing of languages that do not use spaces to delimit words. Both are trained on 100 languages on Wikipedia, and use $\alpha = 0.7,\ 0.5$ respectively. XLM also uses language embeddings, and incorporates Translation Language Modeling (TLM) as a pretraining technique, which requires parallel data. XLM-R~\cite{conneau2019unsupervised} handles all languages without preprocessing. It decreases $\alpha$ to $0.3$ and switches to the CommonCrawl dataset, which has a more balanced language distribution. It also disposes of language embeddings and TLM. XLM-T~\cite{barbieri2022xlmt} extends XLM-R by finetuning it on tweets to perform multilingual sentiment analysis, as does XLM-EMO~\cite{bianchi2022xlm} for emotion recognition on four emotions.

\subsection{Zero-shot Emotion Recognition}

Very few works explicitly study zero-shot emotion recognition in text with transformer-based models \cite{yin2019benchmarking, plaza2022natural}. They do so by formulating the problem as a \textit{Natural Language Inference} problem, i.e., by creating a different prompt per emotion of interest, and classifying whether each prompt follows from the input sequence (entailment) or not (contradiction). This requires running the model once per emotion, creating a bottleneck for classification. Most similar to ours are earlier approaches that used semantic similarity with emotion word embeddings to classify in a zero-shot manner \cite{sappadla2016using}.

\section{Methodology}
\label{sec:method}

We present the technical details of interest for \textit{Demux} and our simple extension for it to handle clusters of emotions. Let $E = \{e_i: i\in[n]\}$ be the set of emotions and $C = \{C_i: i\in[m]\}$ be some clustering of $E$ s.t. $n \ge m$, $\cup_{i\in[m]} C_i = [n]$ and $\cap_{i\in[m]} C_i = \emptyset.$

\subsection{Demux}

Let $x$ be an input text sequence. \textit{Demux} constructs $x^\prime=\text{``}e_1, e_2, \dots\allowbreak \text{or}\ e_n?\text{''}$ and use a LM $L$ with its corresponding tokenizer $T$:
\begin{equation}
\begin{split}
    \tilde{x} = T(x^\prime, x) = (&[CLS], t_{1, 1}, \dots, t_{1, N_1}, \dots,
    \\ & t_{n, 1}, \dots, t_{n, N_n}, [SEP], x_1, \dots, x_l),
\end{split}
\end{equation}
where $x_i$ are the tokens from $x$, $t_{i, j}$ the $j$-th subtoken of $e_i$, and \verb+[SEP]+ and \verb+[CLS]+ are special tokens of $T$.
$\tilde{x}$ is propagated through $L$ to get $\hat{x} = L(\tilde{x})$, where $\hat{x}$ contains one output embedding corresponding to each input token. We denote the output embedding corresponding to $t_{i,j}$ as $\hat{t}_{i, j}\in\mathbb{R}^d$, where $d$ is the feature dimension of $L$.
Finally, \textit{Demux} averages the embeddings of each emotion's subtokens, and predicts using a 2-layer neural network mapping embeddings to scalars, $\text{NN}: \mathbb{R}^d \rightarrow \mathbb{R}$, followed by sigmoid $\sigma$:
\begin{equation}
    \forall i \in [n],\quad p(e_i|x) = \sigma(\text{NN}(\frac{\sum_{j=1}^{N_i}\hat{t}_{i, j}}{N_i}))).
\end{equation}
Notice that the same NN is applied to all emotions. For emotion clusters, we modify $x^\prime$ to contain all emotions from all clusters. After the forward pass through $L$, we instead aggregate across all emotions of a cluster instead of a single emotion and predict, for each cluster:
\begin{equation}
    \forall i \in [m],\quad p(C_i|x) = \sigma(\text{NN}(\frac{\sum_{j\in C_i}\sum_{k=1}^{N_j}\hat{t}_{j, k}}{\sum_{j\in C_i}N_j}))).
\end{equation}

Moreover, when using multilingual models, we keep all emotions in English to retain the same prompt, $x^\prime$, across all languages.

\subsection{Correlation-aware Regularization}

To provide extra supervision to the model and enhance its correlation awareness between emotions, \textit{Demux} includes a label-correlation regularization loss. This loss takes into account the ground-truth labels for each example in its formulation. Therefore, the emotions are split into two groups, the present and the absent emotions based on annotations $y$, $\mathcal{P}$ and $\mathcal{N}$ respectively. Intra-group relationships are regularized, meaning we only pick pairs of emotions when they are both in $\mathcal{P}$ or both in $\mathcal{N}$. The formulation is:

\begin{equation}
    \begin{split}
        \mathcal{L}_{L, \text{intra}}(y, \hat{y}) = \frac{1}{2} & \left[ \frac{1}{|\mathcal{N}|^2 - |\mathcal{N}|} \sum_{(i, j) \in \mathcal{N}^2}^{i > j} e^{\hat{y}_i + \hat{y}_j}\right. +\\
            & \left. \frac{1}{|\mathcal{P}|^2 - |\mathcal{P}|} \sum_{(i, j) \in \mathcal{P}^2}^{i > j} e^{-\hat{y}_i-\hat{y}_j} \right], \\
    \end{split}
\end{equation}
where $\hat{y}$ is the prediction of the model and subscripts indicate indexing. In this manner, we decrease the distance of pairs of emotions when they have the same gold labels. The denominators simply average the terms. The final loss is a convex combination of the classification and the regularization loss, dictated by hyperparameter $c$:

\begin{equation}
    \mathcal{L} = (1 - c)\mathcal{L}_{BCE} + c \mathcal{L}_{L, \text{intra}}.
\end{equation}

\section{Experiments} \label{sec:exp}

\subsection{Datasets}

We use the publicly available SemEval 2018 Task 1 E-c \cite{mohammad2018semeval}, and private data containing tweets from the French elections of 2017.

SemEval 2018 Task 1 E-c (SemEval E-c) contains tweets annotated for 11 emotions in a multilabel setting, namely \textit{anger, anticipation, disgust, fear, joy, love, optimism, pessimism, sadness, surprise,} and \textit{trust}, in \textit{English, Arabic,} and \textit{Spanish}.
The cardinalities are 6838 training, 886 development and 3259 testing for English, 2278 training, 585 development and 1518 testing  for Arabic, and 3561 training, 679 development and 2854 testing for Spanish.
We have also machine-translated the English subset of the tweets into French.

\begin{table}[t]
    \centering
    \begin{tabular}{lrr}

        Emotion Cluster & Support in FrE-A & Support in FrE-B \\
        \midrule

        Admiration & 589 (14.4\%) & 51 (1.1\%) \\
        Sarcasm & 395 (9.6\%) & 233 (5.1\%) \\
        Anger & 595 (14.5\%) & 279 (6.1\%) \\
        Embarrassment & 182 (4.4\%) & 56 (1.2\%) \\
        Fear & 271 (6.6\%) & 116 (2.5\%) \\
        Joy & 169 (4.1\%) & 40 (0.9\%) \\
        Optimism & 501 (12.2\%) & 350 (7.7\%) \\
        Pride & 539 (13.1\%) & 92 (2\%) \\
        Positive-other & 632 (15.4\%) & 1268 (27.7\%) \\
        Negative-other & 542 (13.2\%) & 1426 (31.2\%) \\
        \midrule
        All tweets & 4102 (100\%) & 4574 (100\%) \\
        
    \end{tabular}
    \caption{Support per emotion cluster for our French election datasets.}
    \label{tab:fr-datasets}
\end{table}


        

The French election dataset contains mostly French tweets annotated for 10 clusters of emotions in a multilabel setting, namely \textit{admiration-love, amusement-sarcasm, anger-hate-contempt-disgust, embarrassment-guilt-shame-sadness, fear-pessimism, joy-happiness, optimism-hope, pride (including national pride),} any \textit{other positive}, and any \textit{other negative} emotions. The formulation dictates that if one emotion in the cluster is present, the cluster is considered present. Only cluster-wise annotations are requested.
We have two separate subsets, collected by third parties using keywords related to prominent politicians, agendas and concerns of the French elections, and annotated independently also by third parties. The first (``FrE-A'') contains 4102 tweets that we randomly split into training, development and test sets with ratios 8:1:1. The second subset (``FrE-B'') contains 4574 tweet that we also randomly split with ratios 8:1:1. In Table \ref{tab:fr-datasets}, we present the support for each class. Annotations for FrE-B are a lot sparser.

\subsection{Implementation Details}

We use Python (v3.7.4), PyTorch (v1.11.0) and the \textit{transformers} library (v4.19.2). We use NVIDIA GeForce GTX 1080 Ti.
Given the difference in writing formality on Twitter, we use Twitter-based in addition to general-purpose models. For the former, XLM-T~\cite{barbieri2022xlmt} is our multilingual model, BERTweet~\cite{nguyen2020bertweet} for English, RoBERTuito~\cite{perez2021robertuito} for Spanish, AraBERT~\cite{antoun2020arabert} for Arabic, and BERTweetFR~\cite{guo2021bertweetfr} for French. For the latter, we use multilingual BERT~\cite{devlin2018bert} as our multilingual model, BERT for English and Arabic, and BETO~\cite{CaneteCFP2020} for Spanish.
We retain the hyperparameters used in \cite{chochlakis2022leveraging}, such as the learning rate and its warmup, early stopping, batch size, the convex combination coefficient $c$ ($\alpha$ in \cite{chochlakis2022leveraging}), and the text preprocessor. During multilingual finetuning on SemEval E-c, we sample from different languages with equal probability ($\alpha=0$ in aforementioned normalization in Section \ref{sec:multilingual-related}). We evaluate using the F1 score for individual emotions, and micro F1, macro F1 and Jaccard score (JS) otherwise. We also used Micro F1 for early stopping in FrE-A because we found JS to be unreliable due to the label distribution. 

\subsection{Knowledge Transfer between Languages}

\textbf{Multilingual Training and Evaluation}\quad First, we examine how feasible and competitive it is to use multilingual emotion recognition models trained and evaluated on a mixture of languages. To establish this, we first consider training monolingual and multilingual models, pretrained either on tweets or general-purpose text. For multilingual models, we also consider how training and/or evaluating on multiple languages fares with training and/or evaluating on a single language. For example, we compare performance on the Arabic subset when training and performing early stopping solely on Arabic, with the performance when training on all languages simultaneously, but still performing early stopping on Arabic, and other combinations.

We find that training and evaluating on all languages in SemEval E-c has either only slightly negative or strongly positive influence on accuracy for higher-resource languages. Our dev set results are presented in Table \ref{tab:multi-v-mono-v-twitter}. It becomes immediately obvious that models trained on Twitter comfortably outperform their general-purpose alternatives (first and the second row, and third and final row). We also observe that using a dev set of multiple languages not only does not hurt performance, but actually achieves positive knowledge transfer for Spanish, achieving the best JS overall. The monolingual alternatives perform favorably in English and Arabic, with the increase in the former being relatively minor. Overall, since our ultimate goal is to transfer to French tweets, and given the competent or superior performance for Latin-based languages, we do adopt the multilingual model trained and evaluated on multilingual data for pretraining.

\begin{table*}[t]
    \centering
    \begin{tabular}{c@{\hspace{1.50\tabcolsep}}c@{\hspace{1.50\tabcolsep}}c@{\hspace{ 1.50\tabcolsep}}cccc}

        &&&& \multicolumn{3}{c}{\textbf{JS}} \\
        \cmidrule(lr){5-7}
        \multicolumn{4}{c}{\textbf{Setting}} & En & Es & Ar \\

        \cmidrule{1-4} \cmidrule(lr){5-7}

        & Monolingual models & & & \rescell{61.0}{0.3} & \rescell{53.2}{0.5} & \rescell{55.1}{0.3} \\
        Twitter-based & Monolingual models & &  & \rescell{\mathbf{62.0}}{0.4} & \rescell{55.6}{0.3} & \rescell{\mathbf{61.6}}{0.5} \\
        & Multilingual models & w/ Multilingual Training & \& Evaluation & \rescell{58.4}{1.0} & \rescell{50.3}{0.4} & \rescell{49.2}{0.0} \\
        Twitter-based & Multilingual models && & \rescell{61.6}{0.1} & \rescell{56.8}{1.0} & \rescell{56.5}{0.6} \\
        Twitter-based & Multilingual models & w/ Multilingual Training & & \rescell{61.3}{0.4} & \rescell{56.5}{0.6} & \rescell{57.9}{1.0} \\
        Twitter-based & Multilingual models & w/ Multilingual Training & \& Evaluation & \rescell{61.3}{0.2} & \rescell{\mathbf{58.0}}{0.7} & \rescell{57.7}{0.4} \\
        
    \end{tabular}
    \caption{Comparing Jaccard scores in \textit{SemEval 2018 Task 1 E-c}. The variables we consider are: monolingual or multilingual models, Twitter-based or general-purpose models, monolingual or multilingual training, where the training set is comprised of one or a mixture of all languages, and monolingual or multilingual evaluation, where the evaluation set is comprised of one or a mixture of all languages.}
    \label{tab:multi-v-mono-v-twitter}
\end{table*}

\textbf{Transfer to New Languages}\quad In trying to establish how emotion recognition knowledge is transferred to new languages, we conducted experiments with one SemEval E-c language left out during training. Results are shown in Table \ref{tab:lang-zeroshot}. We notice a drop in performance, with all models performing roughly equivalently across all metrics on the new language notwithstanding original performance. In detail, all metrics drop by around $25\%$ in Arabic, $<22\%$ for Spanish, while the drop in English ranges from $18\%$ to $32\%$. Nonetheless, emotion recognition in the new language occurs at a competent level, picking up clear signals despite the noise from the language switch, rendering multilingual emotion recognition models capable of being used with new languages. 

\begin{table}[t]
    \centering
    \begin{tabular}{ccccccc}

        &&&& \multicolumn{3}{c}{SemEval 2018 Task 1 E-c} \\
        \cmidrule(lr){5-7}
        \multicolumn{3}{c}{Train langs} & Eval langs & Mic-F1 & Mac-F1 & JS \\
        \toprule
        En & Es & & Ar & \rescell{55.8}{0.9} & \rescell{42.4}{2.7} & \rescell{43.3}{1.3} \\
        En & Es & Ar & Ar & \rescell{70.2}{0.8} & \rescell{57.4}{1.4} & \rescell{57.9}{1.0} \\
        \midrule
        En & & Ar & Es & \rescell{55.9}{0.7} & \rescell{45.1}{1.6} & \rescell{44.0}{0.9} \\
        En & Es & Ar & Es & \rescell{65.1}{0.6} & \rescell{54.7}{0.8} & \rescell{56.5}{0.6} \\
        \midrule
        & Es & Ar & En & \rescell{54.6}{0.6} & \rescell{47.0}{1.3} & \rescell{41.9}{0.7} \\
        En & Es & Ar & En & \rescell{72.3}{0.3} & \rescell{57.6}{2.0} & \rescell{61.3}{0.4} \\
    \end{tabular}
    \caption{Leave-one-language-out experiments.}
    \label{tab:lang-zeroshot}
\end{table}

\subsection{Knowledge Transfer to New Emotions}

We also assess \textit{Demux}'s ability to perform zero-shot emotion recognition by excluding emotions from SemEval E-c, one at a time. We can predict unseen emotions since the final classifier maps embeddings to probabilities, agnostic to specific emotions. We choose \textit{anger}, \textit{joy}, \textit{pessimism}, and \textit{trust} to capture the change in accuracy across a wide spectrum of performance levels. In particular, \textit{joy} and \textit{trust} are the highest and lowest performing emotions, whereas \textit{anger} and \textit{pessimism} have relatively high and low scores, respectively. Results are presented in Table \ref{tab:emotion-zeroshot}. We notice a decrease in performance. However, the model can still predict these unseen emotions, especially those with which the original model was competent at.

To remedy the decrease in performance, we experimented with freezing word embeddings in an effort to retain the relationships between emotions. We examine two alternatives, freezing the word embedding layer altogether, and freezing only the emotion word embeddings (including the novel emotions). Results are also presented in Table \ref{tab:emotion-zeroshot}. We find this decreases performance for the model, indicating it already captured relationships across emotions in its input.

\begin{table}[t]
    \centering
    \begin{tabular}{cccccc}

        && \multicolumn{4}{c}{ SemEval 2018 Task 1 E-c F1} \\
        \cmidrule(lr){3-6}
        ZS & Frozen & Anger & Joy & Pessimism & Trust \\
        \toprule
        \xmark & - & \rescell{78.6}{0.3} & \rescell{86.0}{0.3} & \rescell{40.5}{1.5} & \rescell{9.9}{3.9} \\
        \cmark & - & \rescell{42.5}{11.8} & \rescell{57.8}{3.2} & \rescell{11.7}{0.9} & \rescell{3.6}{3.5} \\
        \cmark & Words & \rescell{25.1}{18.2} & \rescell{51.5}{7.3} & \rescell{15.1}{12.9} & \rescell{1.7}{2.0} \\
        \cmark & Emos & \rescell{21.5}{20.8} & \rescell{36.2}{14.3} & \rescell{4.3}{4.3} & \rescell{3.7}{4.6} \\
    \end{tabular}
    \caption{Zero-shot performance of unseen emotions.}
    \label{tab:emotion-zeroshot}
\end{table}

\begin{table}[t]
    \centering
    \begin{tabular}{cc@{\hspace{0.95\tabcolsep}}c@{\hspace{0.95\tabcolsep}}ccc}
        &&& \textbf{Mic-F1} & \textbf{Mac-F1} & \textbf{JS} \\
        \cmidrule(r){4-6}
        & \textbf{Pretrained} & \textbf{Finetuned} & \multicolumn{3}{c}{FrE-A} \\
        \toprule
        
        &\multicolumn{2}{c}{Most frequent} & $0$ & $0$ & $24.9$ \\
        &\multicolumn{2}{c}{Uni. Random} & \rescell{17.6}{0.4} & \rescell{18.0}{0.4} & \rescell{10.0}{0.2} \\

        \midrule
        
        \parbox[t]{2mm}{\multirow{3}{*}{\rotatebox[origin=c]{90}{XLM-T}}} & \cmark & \xmark & \rescell{21.0}{1.2} & \rescell{15.8}{2.0} & \rescell{26.6}{2.2} \\  
        &\xmark & \cmark & \rescell{28.9}{1.4} & \rescell{25.7}{0.7} & \rescell{\mathbf{29.8}}{1.7} \\
        &\cmark & \cmark & \rescell{\mathbf{30.5}}{2.5} & \rescell{\mathbf{26.5}}{5.2} & \rescell{29.4}{2.3} \\
        
        \midrule
        
        \parbox[t]{2mm}{\multirow{3}{*}{\rotatebox[origin=c]{90}{\makecell{BERT\\TweetFr}}}} & \cmark & \xmark & \rescell{24.8}{0.7} & \rescell{18.5}{1.4} & \rescell{15.2}{0.6} \\
        &\xmark & \cmark & \rescell{27.0}{12.5} & \rescell{24.9}{12.8} & \rescell{29.6}{2.6} \\
        &\cmark & \cmark & \rescell{\mathbf{34.3}}{1.1} & \rescell{\mathbf{31.4}}{1.7} & \rescell{\mathbf{31.8}}{1.0} \\
        
        \cmidrule(r){4-6}
        &&& \multicolumn{3}{c}{FrE-B} \\
        \cmidrule(r){4-6}
        
        &\multicolumn{2}{c}{Most frequent} & $0$ & $0$ & $35.0$ \\
        &\multicolumn{2}{c}{Uni. Random} & \rescell{15.0}{0.7} & \rescell{12.3}{0.7} & \rescell{8.3}{0.4} \\

        \midrule
        
        \parbox[t]{2mm}{\multirow{5}{*}{\rotatebox[origin=c]{90}{XLM-T}}} & \cmark & \xmark & \rescell{18.3}{3.7} & \rescell{11.5}{1.9} & \rescell{29.4}{4.6} \\
        &\xmark & \cmark & \rescell{55.3}{6.5} & \rescell{19.3}{9.9} & \rescell{55.3}{6.5} \\
        &\cmark & \cmark & \rescell{\mathbf{62.7}}{0.9} & \rescell{\mathbf{44.4}}{2.8} & \rescell{\mathbf{59.3}}{0.8} \\ 
        
        \cmidrule{2-6}
        &\multicolumn{1}{l}{* \phantom{11} \cmark} & \xmark & \rescell{15.9}{2.6} & \rescell{11.3}{1.0} & \rescell{14.8}{3.3} \\
        &\multicolumn{1}{l}{* \phantom{11} \cmark} & \cmark & \rescell{61.9}{1.2} & \rescell{39.6}{9.6} & \rescell{58.4}{0.8} \\

        \midrule

        \parbox[t]{2mm}{\multirow{3}{*}{\rotatebox[origin=c]{90}{\makecell{BERT\\TweetFr}}}} & \cmark & \xmark & \rescell{27.9}{1.9} & \rescell{15.9}{1.1} & \rescell{15.3}{1.4} \\
        &\xmark & \cmark & \rescell{63.8}{1.4} & \rescell{29.7}{7.6} & \rescell{60.2}{1.7} \\
        &\cmark & \cmark & \rescell{\mathbf{66.9}}{0.5} & \rescell{\mathbf{40.3}}{6.0} & \rescell{\mathbf{63.2}}{1.3} \\

    \end{tabular}
    \caption{Performance in French election data when the models pretrain on \textit{SemEval E-c} and/or finetune on the corresponding dataset. *: maximum probability instead of pooling embeddings in Demux.}
    \label{tab:clusters}
\end{table}

\subsection{Knowledge Transfer to New Annotation Format}

Finally, we evaluate if the model can successfully transfer knowledge to a new annotation format. In particular, we transfer from SemEval E-c to FrE-a and FrE-B. The former contains tweets in English, Spanish and Arabic, and annotations for emotions. The latter contain French tweets annotated for clusters. Additionally, the emotions of the latter are not a subset of the former. Therefore, we expect to observe compounding effects from the multiple changes in the setting. To address the language switch, we also use the French translations to train BERTweetFr as a monolingual parallel to \mbox{XLM-T}.

We also show two simple baselines, one that predicts the most frequent label for each emotion, and another that predicts uniformly randomly. Moreover, our experiments include an alternative to averaging emotion embeddings with \textit{Demux}, where we instead select the maximum predicted probability across emotions in a cluster.

Results are presented in Table \ref{tab:clusters}. We consider three different settings for each model and each dataset in order to study how training on SemEval E-c (\textit{Pretrained}) and training on the French election datasets (\textit{Finetuned}) affects performance. In the first row per model and dataset, we see the zero-shot performance on the dataset. The second shows the performance of a model only trained on the dataset. The last row shows performance from a model first trained on SemEval E-c and then on the corresponding dataset.

For FrE-A, zero-shot performance of XLM-T is close to either finetuned performance, indicating the model can successfully transfer despite the complete change of environment. Performance increases for both when training on French data, and when SemEval E-c is also included. The monolingual model performs favorably only when fully supervised, which indicates that translations are not ideal for zero-shot transfer, and overall proves the ability of multilingual models to transfer knowledge to a new language.

Performance is evidently better in FrE-B. Again, XLM-T performs better than random and favorably to BERTweetFr in the zero-shot setting. Finetuning improvements are significant for both models. Lastly, the alternative pooling performs worse than our proposed one.
This speaks to the subjectivity of the annotations, allowing operations on emotion embeddings but not the predictions themselves.  

\section{Conclusion}
\label{sec:conc}

In this work, we study how to transfer emotion recognition knowledge to different languages, different emotions, and different annotation formats. We find that multilingual models have the capacity to transfer that kind of knowledge sufficiently well. In order to transfer knowledge between emotions, we leverage \textit{Demux}'s transferability between emotions through word-level associations. We see that the model also inherently performs zero-shot emotion recognition without the need for further changes. Finally, we modify \textit{Demux} to perform aggregation operations on its label embeddings, and show this can transfer knowledge to novel annotation formats, such as clusters of emotions, even in conjunction with the presence of novel emotions and in a different language. We show that multilingual models pretrained on other languages perform favorably in the zero-shot setting to native models pretrained on machine translations.





\bibliographystyle{acm}
\bibliography{refs}

\end{document}